# Adaptive Learning Rule for Hardware-based Deep Neural Networks Using Electronic Synapse Devices

**Suhwan Lim, Jong-Ho Bae, Jai-Ho Eum, Sungtae Lee, Chul-Heung Kim, Dongseok Kwon, Byung-Gook Park, Jong-Ho Lee\***

Department of Electrical and Computer Engineering, Inter-University Semiconductor Research Center (ISRC), Seoul National University, Seoul, South Korea

**\* Correspondence:**
jhl@snu.ac.kr

**Abstract**

In this paper, we propose a learning rule based on a back-propagation (BP) algorithm that can be applied to a hardware-based deep neural network (HW-DNN) using electronic devices that exhibit discrete and limited conductance characteristics. This adaptive learning rule, which enables forward, backward propagation, as well as weight updates in hardware, is helpful during the implementation of power-efficient and high-speed deep neural networks. In simulations using a three-layer perceptron network, we evaluate the learning performance according to various conductance responses of electronic synapse devices and weight-updating methods. It is shown that the learning accuracy is comparable to that obtained when using a software-based BP algorithm when the electronic synapse device has a linear conductance response with a high dynamic range. Furthermore, the proposed unidirectional weight-updating method is suitable for electronic synapse devices which have nonlinear and finite conductance responses. Because this weight-updating method can compensate the demerit of asymmetric weight updates, we can obtain better accuracy compared to other methods. This adaptive learning rule, which can be applied to full hardware implementation, can also compensate the degradation of learning accuracy due to the probable device-to-device variation in an actual electronic synapse device.

## 1 Introduction

Conventional computing systems which are based on CMOS logic and the von Neumann architecture are powerful for well-defined mathematical problems. However, problems in the real world cannot be efficiently solved using these computing systems. Therefore, biologically inspired neuromorphic systems have emerged as an attractive field of research (Poon and Zhou, 2011). Recently, several types of emerging electronic synapse devices such as phase change memory (PCRAM) (Suri et al., 2011; Wright et al., 2011; Kuzum et al., 2012), resistive change memory (RRAM) (Jo et al., 2010; Ohno et al., 2011; Wu et al., 2012; Yu et al., 2012), ferroelectric devices (Chanthbouala et al., 2012), and FET-based devices (Diorio et al., 1996; Ziegler et al., 2012; Kim et al., 2016) have been proposed to mimic biological synapses. Although most of these works have focused on spike-timing-dependent-plasticity (STDP, (Bi and Poo, 1998)) learning algorithm (Kuzum et al., 2013), the learning performance using STDP is still in its early stage (Burr et al., 2014; Nair and Dudek, 2015). Unlike the approach in which STDP is used, electronic synapse devices (Burr et al., 2014; Prezioso et al., 2015; Merrikh-Bayat et al., 2015) can also be applied to deep neural networks (DNNs) with well-studied back-propagation (BP) algorithms (Rumelhart et al., 1986). The vector-by-matrix multiplication of forward and backward propagations accounts for a large portion of the computational tasks of a software-based deep neural network (SW-DNNs) and thus is the main cause of enormous amounts of power consumed. However, when the input signal and the weight are replaced by the input voltage and the conductance of the electronic synapse device, respectively, the output signal is simply expressed as the current flowing from the electronic synapse device array. Therefore, the use of an electronic synapse device can greatly reduce power consumption and improve the speed (Merkel et al., 2016). This approach represents one type of hardware-based deep neural network (HW-DNN). In this approach, all computations including propagation should be implemented by hardware (HW) using electronic circuits compatible with the electronic synapse device array. If only forward propagation is performed by HW, backward propagation and weight updates based on a BP algorithm require software (SW) support, which may offset any gains stemming from the use of the electronic synapse device array. In other words, when the BP algorithm is calculated by a SW operation, the communication between the SW and the electronic synapse device array can be the dominant bottleneck, greatly reducing the advantages of HW-DNNs. That is, all computations of SW-based deep neural networks, including forward propagation, backward propagation, and weight updating, should be replaced by calculations using electronic synapse device arrays and electronic circuits. The implementation of full HW-DNNs is challenging owing to the complex computations of activation functions and their derivatives (Neftci et al., 2017). When the HW includes electronic synapse devices, HW implementation is much more complicated because the weights should be represented using the conductance of a non-ideal synapse device. Therefore, appropriate electronic synapse devices and electronic circuits compatible with the electronic synapse devices are needed. Moreover, an adaptive learning rule which especially enables hardware implementation using electronic synapse devices and electronic circuits is necessary.

The electronic synapse devices for HW-DNNs should have the following characteristics: an extremely operating low power, high scalability (density), high repeatability and reliability, the ability to combine storage and computations (Kuzum et al., 2013), and linear and symmetric conductance responses (Burr et al., 2014).

In large neural networks based on electronic synapse devices, large synapse array can increase total occupied area and power consumption. Therefore, each electronic synapse device should be scalable and operated in a very low current regime. Furthermore, the electronic synapse device should offer good repeatability, reliability, and uniformity between devices. Otherwise, neural networks (NNs) will diverge because the weights cannot be accurately updated with reference to the target value. If electronic synapse devices have nonlinear and asymmetric conductance response, NNs will no longer learn or diverge because the learning rate is too small or too large depending on the conductance states. In other words, the linear and symmetric conductance response of the electronic synapse device ensures a constant learning rate and convergence of NNs (Burr et al., 2014). Note that a nonlinear conductance response can also degrade the STDP learning algorithm (Merrikh-Bayat et al., 2013). There have been attempts to obtain linear conductance responses (Jang et al., 2015; Woo et al., 2016; Fuller et al., 2017; Bae et al., 2017), and such attempts should not burden the external circuit. Thus, basically, electronic synapse devices should have linear, symmetric, and repeatable conductance responses for given identical pulses.

Electronic circuits should control the electronic synapse device array and efficiently provide an activation function (Gokmen and Vlasov, 2016; Fumarola et al., 2016). Analog VLSI circuits using CMOS FETs have been proposed to achieve a HW-DNN (Binas et al., 2016). In this paper, the integration density of synapses is quite low, because the analog circuits occupy a large area. To increase the integration density of synapses, emerging electronic synapse devices should be adopted. However, electronic synapse devices mentioned above have a nonlinear and finite conductance response, and have a variation in conductance between the devices fabricated on the same substrate. Thus, electronic circuits are required to control the electronic synapse array appropriately depending on the device characteristics. These control circuits are able to apply the required voltage to the electronic synapse device to update the weight to the target value. The activation function must be designed according to the device characteristics. These electronic circuits should be designed efficiently so as not to offset the benefits of the electronic synapse devices.

Above all, learning rules based on a BP algorithm must be applicable to the HW used, including the electronic synapse device array and the electronic circuits. In HW-DNNs which use electronic synapse devices, unlike SW-DNNs, the weights are represented by their conductance values, meaning that the weights have discrete and limited values. These characteristics of the weights can degrade the learning accuracy because it becomes impossible to update the weights precisely to the target values. Therefore, we need an appropriate learning rule to minimize the degradation of the learning accuracy while maintaining the advantages of the electronic synapse device array.

In this paper, we propose an adaptive learning rule based on a BP algorithm for a HW-DNN and its corresponding NNs. First, a SW-DNN is introduced and it is converted to a HW-DNN. In addition, weight-updating methods for HW-DNNs are proposed. We then explain the design of the NNs, which uses an electronic synapse array. Lastly, the learning accuracy of the designed NNs is evaluated with respect to various device characteristics and weight-updating methods. All results in this work are obtained by simulation based on the learning rule for HW-DNNs.

## 2 Materials and Methods

### 2.1 Software- and Hardware-based DNNs

A SW-DNN consists of the input layer, the hidden layers, and the output layer. Each node of the layer is known as a neuron, and a node connection between adjacent layers is called a synapse. The strength of a synapse is the synaptic weight, or simply the weight. The goal of learning in NNs is to specify the hypothesis by changing the weights in a way that minimizes training and generalization errors. The method by which the weight is changed to the target value is mathematically well studied in reference to BP algorithms (Rumelhart et al., 1986).

Like SW-DNNs, HW-DNNs also have layers consisting of neurons and synapses. In HW-DNNs, the neurons and the synapses should be appropriately implemented by HW. In this HW-DNNs we focused on, we assume that synapses are implemented by electronic synapse devices, not VLSI circuits. **Table 1** shows the difference between SW-DNNs and HW-DNNs.

Two electronic devices are required to represent a unit synapse, because the weights ($W_{ij}$ for the weight of the synapse between the $i$th neuron in the $l$–1 layer and the $j$th neuron in the $l$ layer) of the unit synapse in NNs should have positive and negative values for learning. Here, the $l$–1 and $l$ layers represent the layers on which pre-neurons and post-neurons are located, respectively. The input signal ($a_i^{(l-1)}$ for the $i$th neuron in the $l$–1 layer) and the weight ($W_{ij}$) can be represented by the applied voltage ($V_i^{(l-1)}$) and the conductance difference of the unit synapse ($G_{ij}^+ - G_{ij}^-$), respectively. Because the conductance of each electronic device can only be positive, it is necessary to subtract the output current from a pair of synapses to implement the positive and negative weights ($W_{ij} = G_{ij}^+ -$ $G_{ij}^-$) represented by the conductance differences. Subsequently, $(G_{ij}^+ - G_{ij}^-)V_i$ from a unit synapse (a pair of electronic synapse devices) is expressed as the current. By connecting all unit synapses to the $j^{th}$ neuron in the $l$ layer, the currents from each unit synapse are summed. This current is determined by the weighted sum value ($s_j^{(l)}$ for the $j^{th}$ neuron in the $l$ layer). The weighted sum value ($s_j^{(l)}$) is then converted to the input signal of the next layer ($a_j^{(l)}$) using an activation function ($f$), which is implemented by electronic circuits. Through these sequences, we can perform the forward propagation computation using HW. However, if the activation function ($f$) is a sigmoid or a hyperbolic tangent, implementing the activation function in HW is not simple because the derivative value of the activation function has a continuous value. Therefore, for ease of HW implementation, we used a hard-sigmoid function with derivative values of either 0 or $1/(2c)$, as shown in **Figure 1**.

The activated value ($f(s_j)$) equals 0 when the weighted sum value ($s_j$) is less than $-c$ and equals 1 when the weighted sum value ($s_j$) is greater than $c$. Otherwise, the activated value ($f(s_j)$) represents a linear function ($s_j/(2c)$), where $c$ is a positive constant. The parameter $c$ must be chosen by considering the training data sets and the number of electronic synapse devices connected to a post-neuron. Thus, this activated value ($f(s_j)$) has an output value (0 or 1 or $s_j/(2c)$) and an appropriate derivative value (0 or $1/(2c)$) for HW simplicity, as shown in **Figure 1**. Similarly, in SW-DNNs, linear activation functions such as a rectified linear unit (ReLU) and a hard-sigmoid function are commonly used to avoid the vanishing gradient problem. If the derivative of activation function is less than 1, the gradients of the front layers decrease as the number of layers increases. At this point, the weights in the front layers cannot be updated.

In addition to the forward propagation, to compute the backward propagation, the backward-weighted sum should be ascertained after the weight matrices are transposed. In other words, the post-neurons during forward propagation must perform the function of the pre-neurons in backward propagation, and vice versa. That is, the electronic synapse device array should be transposable to implement backward propagation. The input signal in the backward direction ($\delta_j^{(l)}$ for the $i^{th}$ neuron in the $l$ layer) and the weight ($W_{ij}$) are then determined by applied voltage ($V_j^{(l)}$) and the conductance difference in the unit synapse ($G_{ij}^+ - G_{ij}^-$), respectively. In this way, the backward-weighted sum ($\sum_j^M W_{ij}\delta_j^{(l)}$) can be determined by connecting all unit synapses to the $i^{th}$ neuron in the $l-1$ layer. In addition, we should obtain the delta value of the $i^{th}$ neuron in the $l-1$ layer ($\delta_i^{(l-1)}$) by multiplying the derivative value of the activation function ($f'(s_i^{(l-1)})$) by the backward-weighted sum value ($\sum_j^M W_{ij}\delta_j^{(l)}$). Given that the hard-sigmoid function is used as the activation function, the derivative value is 0 or $1/(2c)$. In this case, we can replace the value of $1/(2c)$ with 1 without any logical error, indicating that the delta value of the $i^{th}$ neuron in the $l-1$ layer ($\delta_i^{(l-1)}$) is identical to the backward-weighted sum value ($\sum_j^M W_{ij}\delta_j^{(l)}$) or 0, for reasons which will be explained later in the paper. Most important is how much the weight change. In SW-DNNs, the expression of $\Delta W_{ij}$ is the product of the learning rate ($\eta$), the delta value of the post-neuron ($\delta_j^{(l)}$), and the activated value of the pre-neuron ($f(s_i^{(l-1)})$). However, in HW-DNNs, the weight, which is expressed by the conductance of the electronic synapse device, shows discrete and limited characteristics, indicating that the learning rate depends on the conductance step size. As the number of steps from minimum conductance to maximum conductance is limited in practical electronic devices, the learning rate ($\eta$) is not needed because the weight of a unit synapse is changed by one small conductance step size (one step) per iteration. In this case, it becomes necessary to consider the sign of the delta value of the post-neuron ($\text{sgn}(\delta_j^{(l)})$) and the sign of the activated value of the pre-neuron ($\text{sgn}(f(s_i^{(l-1)}))$). As the activated value of the pre-neuron is 0 or has a positive value, the sign of $\Delta W_{ij}$ ($\text{sgn}(\Delta W_{ij})$) is 0 or the sign of the delta value of the post-neuron ($\text{sgn}(\delta_j^{(l)})$). Therefore, we can replace the value of $1/(2c)$ with 1 without affecting the sign of the delta value of the post-neuron ($\text{sgn}(\delta_j^{(l)})$). This eliminates the vanishing gradient problem and allows us to expand the NNs more deeply. Above all, calculations using these simple gradient values can be implemented without difficulty on the HW. Identical pulses should then be generated by an electronic circuit according to $\text{sgn}(\Delta W_{ij})$ and applied to the electronic synapse devices to change the conductance.

Lastly, updating the weight of each training sample, which is a type of online learning, can easily be implemented on HW. If we adopt batch learning, computation in which the delta value of the post-neuron ($\delta_j^{(l)}$) is multiplied by the activated value of the pre-neuron ($f(s_i^{(l-1)})$) is an additional vector-by-matrix multiplication. This vector-by-matrix multiplication is not a product of the input signal and the weight and therefore cannot be performed by an electronic synapse device array. Consequently, to make hardware implementation easier, online learning is appropriate.

Based on what has been presented thus far, the proposed BP algorithm is a very promising candidate for implementing HW-DNNs without sacrificing the benefits of electronic synapse device arrays.

## 2.2 Weight-updating methods

As noted in the previous section, HW-DNNs using electronic synapse devices need a modified weight-updating method, unlike a SW-based BP algorithm. It is assumed that the weight is updated by changing the conductance of the electronic devices ($G^+$ or $G^-$) by one step. If multiple pulses are required to be applied to an electronic synapse device to arrive at a target conductance, it becomes necessary to check the current conductance of the device and determine the number of pulses required to obtain the target conductance. Because these procedures are needed for all electronic synapse devices, the external circuit can be a major burden. Furthermore, it will be much more difficult if the electronic synapse devices have nonlinear conductance responses. In other words, the use of multiple pulses for precise weight updates is impractical in actual electronic devices (Nair and Dudek, 2015). Therefore, we propose a weight-updating method based on a BP algorithm for HW-DNNs in which the amount of the weight change (equivalently, the learning rate in SW-DNNs) is determined by the amount of the conductance change of the electronic synapse devices.

We only need to consider the sign of the delta value of the post-neuron ($\text{sgn}(\delta_j^{(l)})$) and the sign of the activated value of the pre-neuron ($\text{sgn}(f(s_i^{(l-1)}))$) to determine whether the weight is changed or not. Because the activated value is positive, except when it is 0, only the sign of the delta value of the post-neuron ($\text{sgn}(\delta_j^{(l)})$) is required to determine whether the weight is increased or decreased. Although this HW-based BP algorithm can be less accurate than a SW-based BP algorithm, relieving the burden on the external circuitry can be more helpful for power-efficient HW-DNNs.

After determining whether the weight is to be increased or decreased depending on the sign of the delta value of the post-neuron ($\text{sgn}(\delta_j^{(l)})$), the method by which the weight is changed by changing the conductance of the synapse device is important. As two identical electronic devices ($G^+$ and $G^-$) are necessary for a unit synapse device, the change in the weight can be expressed by the change in the conductance of each device. In other words, we can obtain the same result of increasing the weight by only increasing $G^+$ or decreasing $G^-$. However, as this is slightly redundant work, we choose only to increase the conductance to change the weight. The detailed weight-updating methods are described in **Figure 2**.

If $G^+ < G_{max}$, and $\Delta W > 0$ is necessary, $G^+$ should be increased by one step. Likewise, $G^-$ should be increased under a condition of $G^- < G_{max}$, with $\Delta W < 0$. We simply need to determine whether $G^+$ or $G^-$ is to be increased to update the weight. However, when $G^+$ or $G^-$ reaches $G_{max}$, there can be three different weight-updating methods because we cannot update the weight any more by increasing the conductance of $G^+$ or $G^-$. As an example, a case when $G^+ = G_{max}$ and $\Delta W > 0$ is given, with both $G^+$ and $G^-$ initialized, with a subsequent increase in $G^+$ a possible solution according to the update method described in **Figure 2A**. This update method has been reported (Bichler et al., 2012). For the same case, it is also possible to initialize $G^-$, followed by increasing $G^-$ to a conductance level lower by one step than the previous value, as shown in **Figure 2B**. This method is proposed in this work.

These two weight-updating methods can be implemented by electronic synapse devices which have unidirectional conductance responses and which can be reset. Although the increase in the iterative conductance after initializing can be tedious, it is not a frequent occurrence and is therefore not a major burden. In a bidirectional device, the one-step depression of $G^-$ is equivalent to an increase in the weight by one step, which has a degree of freedom compared to that of a unidirectional device (**Figure 2C**). Additionally, it is necessary to initialize both $G^+$ and $G^-$ when they reach $G_{max}$, because a lower conductance level results in less power consumption (**Figure 2D**). If the target $W$ is zero, either $G^+ = G^- = G_{min}$ or $G^+ = G^- = G_{max}$ results in the same zero $W$. We can then significantly reduce the power consumption by taking, for example, $G^+ = G^- = G_{min}$ instead of $G^+ = G^- = G_{max}$. Because there are a great many synapses in arrays for large neural networks, high conductance of electronic synapse devices can bring about high power consumption. For better understanding, the three weight-updating methods in **Figures 2A, B, and C** are expressed as in the following pseudocode. In this code, $G^+ = G_{max}$ and $\Delta W > 0$ are assumed as a case.

  **function** weight-updating method a
  $W = G^+ - G^-$
  $G^+ \leftarrow G_{min}, \; G^- \leftarrow G_{min}$
    **while** $G^+ - G^- < W$
      $G^+ \leftarrow G^+ + \Delta G$
    **end while**
  **end function**

  **function** weight-updating method b
  $W = G^+ - G^-$
  $G^- \leftarrow G_{min}$
    **while** $G^+ - (G^- + \Delta G) > W$
      $G^- \leftarrow G^- + \Delta G$
    **end while**
  **end function**

  **function** weight-updating method c
  $G^- \leftarrow G^- - \Delta G$
  **end function**

Because these update methods can control the electronic synapse devices which have discrete and limited conductance responses, they can be applied to HW-DNNs capable of running a BP algorithm.

**2.3 Multi-layer perceptron networks**

To assess the learning feasibility of HW-DNNs using the proposed learning rules, we used a three-layer perceptron network based on an electronic synapse device model, as shown in **Figure 3**. Although there are only three layers used in this case, in the results section it is shown that deeper NNs can also be designed without any difficulty. In order to show that a NN can be implemented using electronic synapse devices, we evaluate the learning accuracy by performing a simulation based on the behavioral model of the electronic synapse device using the MTLAB simulation tool (MATLAB 2016a). Handwritten digits (MNIST, 60000 training sets) are used for supervised learning and 10000 test sets are used to evaluate the classification accuracy (Lecun et al., 1998). Because each image in the MNIST sets has 28×28 pixels, the number of input neurons is 28×28. The three-layer perceptron network consists of 784 input, 200 hidden, and ten output neurons. In addition, one neuron for each layer is added for a bias unit, and it is always driven by an input signal of 1. Because our purpose is to implement this NN using HW, we use the hard-sigmoid function as an activation function for each neuron. The hard-sigmoid function makes HW implementation simple due to the simplicity of its output and derivative values. As described in **Table 1**, we can design the hard-sigmoid function according to the parameter $c$. However, we use the softmax function as an activation function for output neurons. Because SW is used to calculate errors in the hypothesis and target data, we can use the softmax function without considering the difficulties of HW implementation. The softmax function has the advantage of representing the output neuron data as the probability and simplifying the weight-updating method with a cross-entropy loss function. In addition, to simplify the hardware implementation, we apply online learning or a mini-batch method, after which the weights are updated for each mini-batch of the MNIST training examples. If we want to apply batch learning, all neuron data for each training set must be stored in memory. This process becomes a heavy burden if the batch size grows. Therefore, the online learning or the mini-batch approach is suitable for HW-DNNs. **Figure 4** shows that the electronic synapse device array can perform forward and backward vector-by-matrix multiplications. We depict the electronic synapse device element as a simple variable resistance symbol. The electronic synapse device can be a two-terminal device or a three-terminal device. In this work, we use a three-terminal device as an example, as shown in **Figure 4**. We can choose any electronic device whose characteristics are suitable for HW-DNNs.

To implement a negative weight, two identical electronic devices are needed for a unit synapse; hence, the weight is expressed in terms of the conductance difference. To perform forward propagation, the input voltage is applied to the drain (equivalently, one terminal of a two-terminal device) nodes ($D^+$ and $D^-$) and the output current flowing into the $j^{th}$ neuron comes from the source (equivalently, the other terminal of a two-terminal device) nodes ($S_j^+$ and $S_j^-$) by connecting the source nodes of $N$ number of electronic synapse devices. In contrast, the roles of the source and drain in the synapse device array should be reversed to implement backward propagation. Thus, the input voltage is applied to the source nodes ($S^+$ and $S^-$) and the output current flowing into the $i^{th}$ neuron comes from the drain nodes ($D_i^+$ and $D_i^-$) by connecting the drain nodes of $M$ number of electronic synapse devices. We need two separate electrodes for both forward and backward propagation to represent positive and negative weights through the current subtractions. Note that only one (S or D) node can be separated by two ($S^+/S^-$ or $D^+/D^-$) electrodes if we apply positive and negative input voltages. For example, when only the S electrode is separated into $S^+$ and $S^-$, we can undertake backward propagation through the positive voltage at $S^+$ and the negative voltage at $S^-$ without subtracting the currents of $D^+$ and $D^-$. Here, we separate both the S and D nodes into $S^+/S^-$ and D$^+$/D$^-$, respectively, for a more general representation. This transposable vector-by-matrix multiplication by the electronic synapse array is necessary to reduce the power consumption significantly and to improve the computation speed when the forward and backward propagations are performed. In addition, we should change the conductance of the electronic synapse device to a target value according to the weight-updating methods. If a three-terminal device is used as an electronic synapse device, a pulse can be applied to the control gate nodes (CG$^+$ and CG$^-$) so as to change the conductance. In a two-terminal device, a pulse can be applied to one of the two terminals.

## 3  Results

**Figure 5** shows the online learning procedure for the designed NNs. The operation in the shaded boxes can be performed by the electronic synapse device array and its control circuits. SW support is only required to process the input and output data, as indicated by the unshaded boxes. After the MNIST pixel data are normalized from 0 to 1, input neurons are driven by the pixel data converted to a voltage pulse suitable for electronic synapse devices. When applying the input voltage to common input nodes (for example, the drains), the output current is summed automatically by connecting all of the nodes (for example, the sources) to a post-neuron. The electronic circuits for the activation function then convert the current to voltage. As the hard-sigmoid function is used as the activation function of the hidden layers, the parameter $c$ should be appropriately determined. If $c$ is too low, the neural networks may fail to learn due to the discrete and limited weight characteristics. After computing the sign of the delta value of the post-neuron ($\text{sgn}(\delta_j^{(l)})$), the weight can be changed in a stepwise manner using identical voltage pulses. We only need to consider the sign of the delta value of the post-neuron ($\text{sgn}(\delta_j^{(l)})$) to determine whether the weight should be increased or decreased unless $\delta_j^{(l)}$ is zero. Because we deal with online learning, the weight-updating event occurs for each training sample. If the mini-batch size is not 1, the weights are updated for each mini-batch of training samples. After the weights are updated, the classification accuracy is calculated by 10000 test samples of MNIST sets for each iteration. With these learning procedures, we evaluate whether the proposed learning rule including weight-updating methods works well in a three-layer perceptron network. It is also evaluated how the non-ideal characteristics of actual electronic devices, such as the finite resolution, the nonlinearity of the conductance response, and the device variability, can affect the classification accuracy.

To evaluate the accuracy with the non-ideal characteristics of an electronic synapse device, a behavioral model from the literature (Querlioz et al., 2013) shown in Equations (1) and (2) is used.

$$G_p(n+1) = G_p(n) + \alpha_p \exp\left(-\beta_p \frac{G_p(n)-G_{\min}}{G_{\max}-G_{\min}}\right) (1)$$

$$G_d(n+1) = G_d(n) - \alpha_d \exp\left(-\beta_d \frac{G_{\max}-G_d(n)}{G_{\max}-G_{\min}}\right) (2)$$

where $\alpha_p$ and $\beta_p$ are the fitting parameters of the potentiation characteristic. Likewise, $\alpha_d$ and $\beta_d$ are fitting parameters for the depression characteristic. Moreover, $G(n)$ denotes the conductance of the electronic synapse device when $n$ pulses are applied, and $G_{\max}$ and $G_{\min}$ are maximum and minimum conductance, respectively. If $n_{\max}$ pulses are needed to progress from $G_{\min}$ to $G_{\max}$, $n_{\max}$ can be defined as the dynamic range. These equations indicate how the conductance changes with different current conductance values. **Figure 6** shows the normalized conductance according to the number of pulses with respect to the parameter $\beta$. The parameter $\beta$ represents the nonlinearity of the electronic synapse device, and the parameter $\alpha$ determines the step size of the conductance change at given level of nonlinearity ($\beta$). A larger $\beta$ means greater nonlinearity. Commonly used electronic devices as synapse devices such as RRAM and PCRAM exhibit asymmetric nonlinearity with a high rate of increase at low conductance and a highly decreasing rate at high conductance. Therefore, we evaluate only the asymmetric nonlinear shape, as mentioned above. Additionally, **Figure 6** shows that the conductance of each device ranges from the minimum value to the maximum value with $n_{\text{max}}$ number of pulses, known as the dynamic range. Because an actual electronic device has a limited dynamic range, this dynamic range is an important factor related to the learning performance. In this work, we evaluate the accuracy with respect to the weight-updating methods when the nonlinearities are 0, 1, 2, and 3 and the dynamic ranges are 32, 64 and 128 (equivalently 5, 6, and 7 bits).

## 3.1 Evaluation of the learning rule

**Figure 7** shows the classification accuracy according to 10000 test samples of the MNIST. Methods a, b, and c in **Figure 7** denote the weight-updating methods in **Figures 2A–C**, respectively. **Figure 7A–D** represent the accuracies when the nonlinearities ($\beta$) are 0, 1, 2, and 3, respectively. The classification accuracy using a SW-based BP algorithm (continuous weight step) is displayed in the figures with a black line for reference. The red, green, and blue lines in the figures represent the classification accuracies for the HW-based BP algorithm with weight-updating methods a, b, and c, respectively. The final accuracy obtained at 60000 iterations using the SW-based BP is slightly smaller than that of the HW-based BP, because the epoch is just one (60000 iterations); hence, we can obtain greater accuracy with additional repetitions. Here the learning rate is 0.01 in the learning using SW-based BP.

First, the linear conductance responses show the highest and most stable accuracy. For a better analysis of the nonlinearity effect in **Figure 7**, the classification accuracy with respect to the nonlinearity ($\beta$) is shown in **Figure 8** as a parameter of weight-updating method. The dynamic range ($n_{\text{max}}$) is 64, and the mini-batch size is 1 (online learning). Given that the accuracy fluctuates due to the discrete weight steps as shown in **Figure 7**, we depict the error bar obtained with the maximum and minimum accuracy levels in the last 100 iterations. The accuracy when using the SW-based BP is shown as a reference (black circle symbols); it fluctuates slightly as well due to the use of online learning. This figure indicates that the accuracy decreases when the nonlinearity increases. If the conductance response is linear, the conductance step size is identical regardless of the current conductance state, and the variation in $|\Delta W|$ at each weight update is zero. If the nonlinearity is increased, then the variation in $|\Delta W|$ increases, resulting in the degradation of the accuracy. It should be noted that when using method b there is relatively less degradation in accuracy compared to methods a and c. The average accuracy rates for method b are 95.36%, 95.59%, 94.80%, and 93.71%, with nonlinearities ($\beta$) of 0, 1, 2 and 3, respectively. Note the degradation of the accuracy in method a is the most significant among three methods. As the nonlinearity increases, not only the average accuracy degradation but also the accuracy fluctuation becomes serious. The behavior of the accuracy with weight-updating methods will be given in the explanation of **Figure 11**.

Secondly, accuracy increases and stabilizes as the dynamic range of conductance increases. **Figure 9** shows the classification accuracy with respect to the dynamic range ($n_{\text{max}}$) as a parameter of weight-updating method. The nonlinearity ($\beta$) is fixed at 2, and the mini-batch size is 1. A high dynamic range means high resolution (small step) in weight update. Consequently, the higher accuracy can be obtained with the better resolution. Similar to that shown in **Figure 8**, method b shows higher accuracy and smaller accuracy fluctuation compared to methods a and c as the dynamic range ($n_{\text{max}}$) increases from 32 to 128. The average accuracy rates for method b are 92.96%, 94.80%, and 94.71%, for dynamic ranges ($n_{\max}$) of 32, 64, and 128, respectively.

Lastly, as the mini-batch size increases, we find that the accuracy decreases and becomes unstable. **Figure 10** shows the classification accuracy with respect to the mini-batch size and as a parameter of weight-updating method. The nonlinearity ($\beta$) and the dynamic range ($n_{\max}$) are fixed at 2 and 64, respectively. Accuracy degradation with increasing mini-batch size has been also reported (Burr et al., 2014). As noted in the Materials and Methods section, online learning is most appropriate in terms of HW implementation. Therefore, a small mini-batch size leads to better learning accuracy and greater ease of HW implementation. It is interesting that method b provides a higher average accuracy and less accuracy fluctuation than methods a and c. The average accuracy rates for method b are 94.80%, 95.64%, 95.70%, and 80.42%, for mini-batch sizes of 1, 2, 5, and 10, respectively.

Why the accuracy depends on the weight-updating method is explained in **Figure 11**. Note that the nonlinearity evaluated here is asymmetric. That is, it shows a highly increasing rate at a low conductance level and a greatly decreasing rate at a high conductance level. We must consider two cases for a proper analysis. The first case is when $G^+ = G_{\max}$ and $\Delta W > 0$ are given (**Figures 11A–C**). The second case is when $G^+ = G_{\max}$ and $\Delta W < 0$ are given (**Figure 11D**). We assume that both cases have same weights before the update, but the target weights in the two cases are in the opposite directions. The weights in the first and second cases should be increased and decreased, respectively.

Applying method a in the first case, both $G^+$ and $G^-$ are initialized and then $G^+$ increases to the target weight in the low conductance region, resulting in a relatively high increasing rate of the weight (**Figure 11A**). However, a low decrease in the weight rate is obtained in the second case (**Figure 11D**). Thus, it is clear that the asymmetry between the weight increase and decrease is high when method a is applied.

Similarly, if method c is applied in the first case, $G^-$ must be decreased rather than initialization taking place, thus resulting in an increase of $W$ (**Figure 11C**). For an asymmetric shape in **Figure 11**, decrease in the rate of the conductance differs from the increase in the rate of the conductance except when the current conductance is in the middle range. This difference is relatively high in the high $G^-$ region. Therefore, we can also confirm that the asymmetry between the weight increase and decrease is significant when method c is applied. However, the accuracy obtained using method c is better than the accuracy obtained using method a. In method c, when $G^-$ decreases, $G^-$ gains a new conductance value that is not on the increase curve. Better accuracy can be achieved by effectively providing more dense steps.

If method b is applied in the first case, $G^-$ must be initialized, and subsequently it increases (**Figure 11B**). Compared to the second case (**Figure 11D**), it can be seen that the difference between the weight increase and decrease is relatively small because the conductance $G$ changes by only one step from the current state. Therefore, we hold that the proposed method b can implement the closest weight change to symmetry. For this reason, the highest and the most stable accuracy rate is obtained when method b is applied.

Because most synapse devices have nonlinear and asymmetric conductance responses, the proposed weight-updating method (b) will be suitable for HW-DNNs. We can say that just the unidirectional conductance response of the electronic synapse device is sufficient when update method b is applied.

Although we used three-layer perceptron networks, the proposed learning rule can be efficiently applied to deeper neural networks. **Figure 12** shows the classification accuracy with respect to the number of hidden layers. When the number of hidden layers is 2, the number of neurons of the first and second hidden layers is 300 and 100, respectively. The number of neurons of the first, second, and third hidden layers is 400, 200, and 100, respectively, in case of the 3 hidden layers. The dynamic range is 64 and the mini-batch size is 1. When the nonlinearity ($\beta$) is 0 and 2, the accuracy is evaluated using the weight-updating method b. The accuracy rates of the SW-based BP and the HW-based BP ($\beta = 0, 2$) are represented by the black circle, and the red and green square symbols, respectively. With two hidden layers, the average accuracy is increased compared to the use of one hidden layer in the neural networks in both the SW-based and HW-based cases. However, the average accuracy and stability are degraded in the neural networks with three hidden layers in one training epoch, but the accuracy rate is improved as training epoch increases. The average accuracies for the SW-based BP are 93.83%, 96.19%, and 91.84%, when the number of hidden layers is 1, 2, and 3, respectively. When the training epoch is 3, the average accuracy of the 3-hidden layer neural networks increases to 95.10% as represented by the black solid circle in **Figure 12**. It seems that a 2-hidden layer neural network is enough for the MNIST training sets, and a neural network with more hidden layers needs regularization methods for performance improvement (Srivastava et al., 2014). The average accuracy rates of the HW-based BP ($\beta = 0$) are 95.36%, 95.85%, and 94.45%, when the number of hidden layers is 1, 2, and 3, respectively. Comparing the results of the SW-based BP with those of the HW-based BP, we find that the results of the HW-based BP are somewhat better. As noted above, if the training epoch is increased, the accuracy rate of the SW-based BP can be improved. After all, the accuracy rate of the HW-based BP is comparable to that of the SW-based BP, and our learning rule also works well for neural networks which have more hidden layers.

## 3.2 On-chip learning robust to device variations

Another important non-ideal characteristic of an electronic synapse device is the device-to-device variation. Even if we can find a physical mechanism that allows electronic synapse devices to have a linear and high dynamic range of the conductance response, we cannot avoid the device-to-device variation which arises during the fabrication process. Therefore, it is needed to investigate the learning accuracy rate in HW-DNNs when there is a variation between devices.

**Figure 13** shows the device-to-device variation in relative conductance scaled by parameter $X$ when the mean conductance is 1 and standard deviation ($\sigma$) is 0.25. We used Gaussian random numbers to represent the parameter $X$. Therefore, the conductance of imperfect electronic synapse devices can vary from $1 - 3\sigma$ to $1 + 3\sigma$, with a probability of 99%.

**Figure 14** shows the classification accuracy with respect to the standard deviation ($\sigma$). Here the nonlinearity ($\beta$), the dynamic range, and the mini-batch size are 2, 64, and 1, respectively. The weight-updating method b is applied. We compared the accuracy rates of on-chip learning and off-chip learning to evaluate the effect of device variations. When we perform off-chip learning, all of the synaptic weights are calculated assuming there is no variation between the devices, with the calculated weights then transferred to an electronic synapse array with the given standard deviation ($\sigma$). In HW-DNNs, it is impractical precisely to map from calculated weights to the electronic conductance while taking into account variations in the entire electronic synapse device array. Whereas when performing on-chip learning, the problems due to the device-to-device variation can be solved. Because HW-DNNs can perform backward propagation using an electronic synapse array, whether the weight should be increased or decreased can be determined based on the current conductance of each imperfect electronic synapse device. Therefore, we can obtain results showing that the accuracy of on-chip learning is not affected by device variations, whereas the accuracy of off-chip learning is significantly reduced when device variations increase. The average accuracy rates of the on-chip learning are 94.92%, 94.81%, and 94.01%, when the standard deviations (σ) are 0, 0.5, and 1, respectively. Whereas, the average accuracies for the off-chip learning are 94.67%, 79.24%, and 57.34%, when the standard deviations (σ) are 0, 0.5, and 1, respectively. The error bars of the off-chip learning are obtained by repeating 10 simulations. In an actual electronic synapse device, a standard deviation (σ) of 0.5 is quite possible as an electronic synapse device should represent multiple analog conductance values with a high dynamic range, unlike digital memory cells having only a 0 or 1 state.

## 4   Discussion

Designing and fabricating an electronic device to act as a synapse in an artificial neural network is a very attractive field of research. In this approach, for power-efficient and high-speed deep neural networks, we should implement forward and backward propagation as well as weight updates using full hardware consisting of electronic synapse devices. However, most electronic devices show non-ideal characteristics such as nonlinearity, asymmetry, and finite conductance responses. Therefore, we need appropriate electronic devices, electronic circuits, and adaptive learning rules for these hardware components.

Here, we applied a learning rule suitable for non-ideal electronic synapse devices which uses a single conductance step to update the weights for each iteration. In an actual electronic device, applying multiple pulses to change the weights by determining the magnitude of the delta value ($\delta$) is a major burden on electronic circuits. It is also impractical to consider variation of an electronic device at each weight update in a very large synapse array. Therefore, it is suitable to determine whether to increase or decrease the weight considering only the sign of the delta value ($\delta$). It means that a learning rate is not needed in HW-DNNs, unlike in SW-DNNs, and it is simply determined by the minimum the conductance steps of electronic devices.

In the results section, we showed how the non-ideal characteristics of electronic synapse devices can affect the learning performance with respect to three different weight-updating methods. If the conductance response is linear, the results of three weight-updating methods are identical, but with a nonlinear conductance response, the results differ due to the asymmetric weight updates. Higher and more stable accuracy can be obtained when the conductance response of the electronic synapse device is as linear as possible. Of course, the higher the dynamic range is, the better the learning performance will be obtained. In addition, increasing the mini-batch size degrades the learning performance, so the online learning is appropriate not only for HW implementation, but also for learning performance. When the conductance response is nonlinear and asymmetric, the variation in $|\Delta W|$ at each weight update is large. It means that for example, the $|\Delta W|$ cannot be zero when the weight at present state is increased and decreased by the same number. However, the proposed unidirectional update method (in this paper method b) can mitigate the decline in the accuracy caused by high nonlinearity, a low dynamic range of the conductance response, and a large mini-batch size. Asymmetric weight updates due to the nonlinearity of the conductance response can be compensated for with the unidirectional update method introduced here (method b), which leads to a low variation in $|\Delta W|$ at each weight update, thus preventing the decline in the accuracy. In actual electronic devices, it is very difficult to ensure that an electronic device is fabricated with linear and symmetric conductance responses. Furthermore, it is difficult to implement repeatable dynamic ranges with more than 100 steps. Therefore, the proposed unidirectional update method (method b) is suitable for actual electronic synapse devices.

Finally, the proposed learning rule which enables on-chip learning is robust to device-to-device variations, which are unavoidable during the fabrication process. When using on-chip learning, the problems due to the device variation is solved; hence, these variations cannot affect the learning accuracy, unlike with off-chip learning. Therefore, the training and the inference should not be separated when configuring HW-DNNs using electronic synapse devices.

As a result of evaluating the effect of the non-ideal characteristics of electronic synapse device on learning performance according to weight-updating methods, we can say that our learning rule is suitable for HW-DNNs which should perform all computations of deep neural networks. Thus, by applying our learning rule, HW-DNNs can accordingly provide high power efficiency, high speeds, and good robustness to device variations.

## 5 Conclusion

Here, we investigated the learning rule suitable for hardware consisting of electronic synapse device array. For ease of hardware implementation, a learning rule including weight-updating methods was recommended. We then evaluated the effects of the conductance response of an electronic synapse device on the learning performance using three different weight-updating methods. If the conductance response is as linear as possible and the dynamic range is as high as possible, better learning performance is obtained. We also confirmed that the proposed unidirectional weight-updating method can mitigate the decline in the accuracy due to non-ideal characteristics of electronic synapse devices. Because most electronic synapse devices have nonlinear, asymmetric, and finite dynamic range conductance responses, the proposed unidirectional weight-updating method is appropriate to obtain the best learning performance. The proposed learning rule enables the implementation of all computations of deep neural networks by a full hardware design, which leads to robustness against device variations. Because device-to-device variations are unavoidable when electronic devices are fabricated in massive arrays, our on-chip learning rule is very helpful when used in HW-DNNs. It is expected that our learning rule will be suitable in implementing HW-DNNs which is efficient with regard to power, capable of high speeds, and immunity to device variations.

## 6 Author Contributions

SHL and JB conceived of the idea and undertook the learning and recognition simulations. JE, STL, CK, and DK supported the simulation and helped with the manuscript preparation step. All authors discussed the results. JL supervised the research.

## 7 Acknowledgements

This work was supported by the KIST Institutional Program (Project No. 2E27330-17-P025) and the Brain Korea 21 Plus Project in 2017.

**Table 1.** Learning rule of software- and hardware-based neural networks

| Target | Software-based | Hardware-based |
|---|---|---|
| Weights $W_{ij}$ | $W_{ij}$ | $G_{ij}^+ - G_{ij}^-$ |
| Forward propagation $s_j^{(l)}$ | $\sum_{i}^{N} W_{ij} a_i^{(l-1)}$ | $a_i^{(l-1)} \to V_i^{(l-1)}$<br>$\sum_{i}^{N} (G_{ij}^+ - G_{ij}^-) V_i^{(l-1)}$ |
| Activated value $a_j^{(l)}$ | $f(s_j^{(l)})$ | $\begin{cases} 0 \text{ if } s_j^{(l)} < -c \\ 1 \text{ if } s_j^{(l)} > c \\ s_j^{(l)} \text{ else} \end{cases}$ |
| Backward propagation $\delta_i^{(l-1)}$ | $\sum_{j}^{M} W_{ij} \delta_j^{(l)} \cdot f'(s_i^{(l-1)})$ | $\delta_j^{(l)} \to V_j^{(l)}$<br>$\begin{cases} \sum_{j}^{M} (G_{ij}^+ - G_{ij}^-) V_j^{(l)} \cdot 0 \\ \sum_{j}^{M} (G_{ij}^+ - G_{ij}^-) V_j^{(l)} \cdot 1 \end{cases}$ |
| Weight updates $\Delta W_{ij}$ | $-\eta \cdot \delta_j^{(l)} \cdot f(s_i^{(l-1)})$ | $\begin{cases} |\Delta G_{ij}^+| \text{ if } \delta_j^{(l)} < 0 \\ -|\Delta G_{ij}^-| \text{ if } \delta_j^{(l)} > 0 \\ 0 \text{ if } f(s_i^{(l-1)}) = 0 \end{cases}$ |

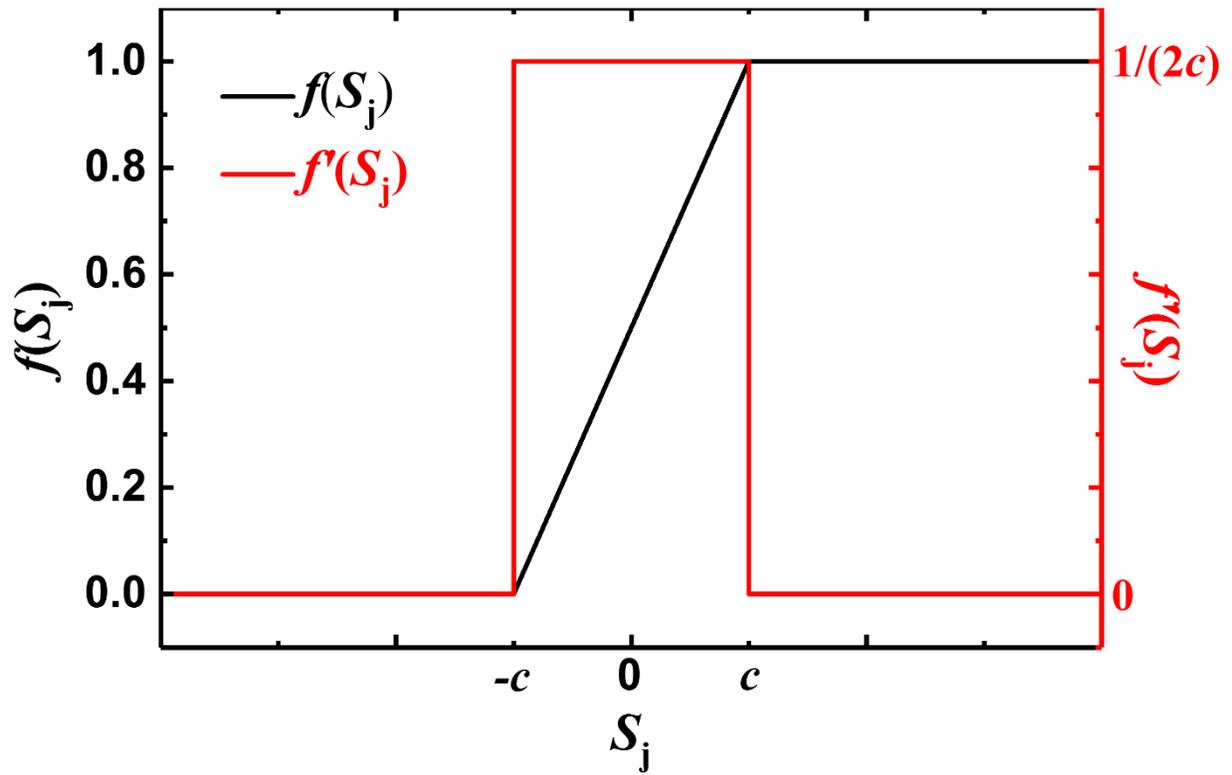

**Figure 1.** Hard-sigmoid activation function and its derivative function

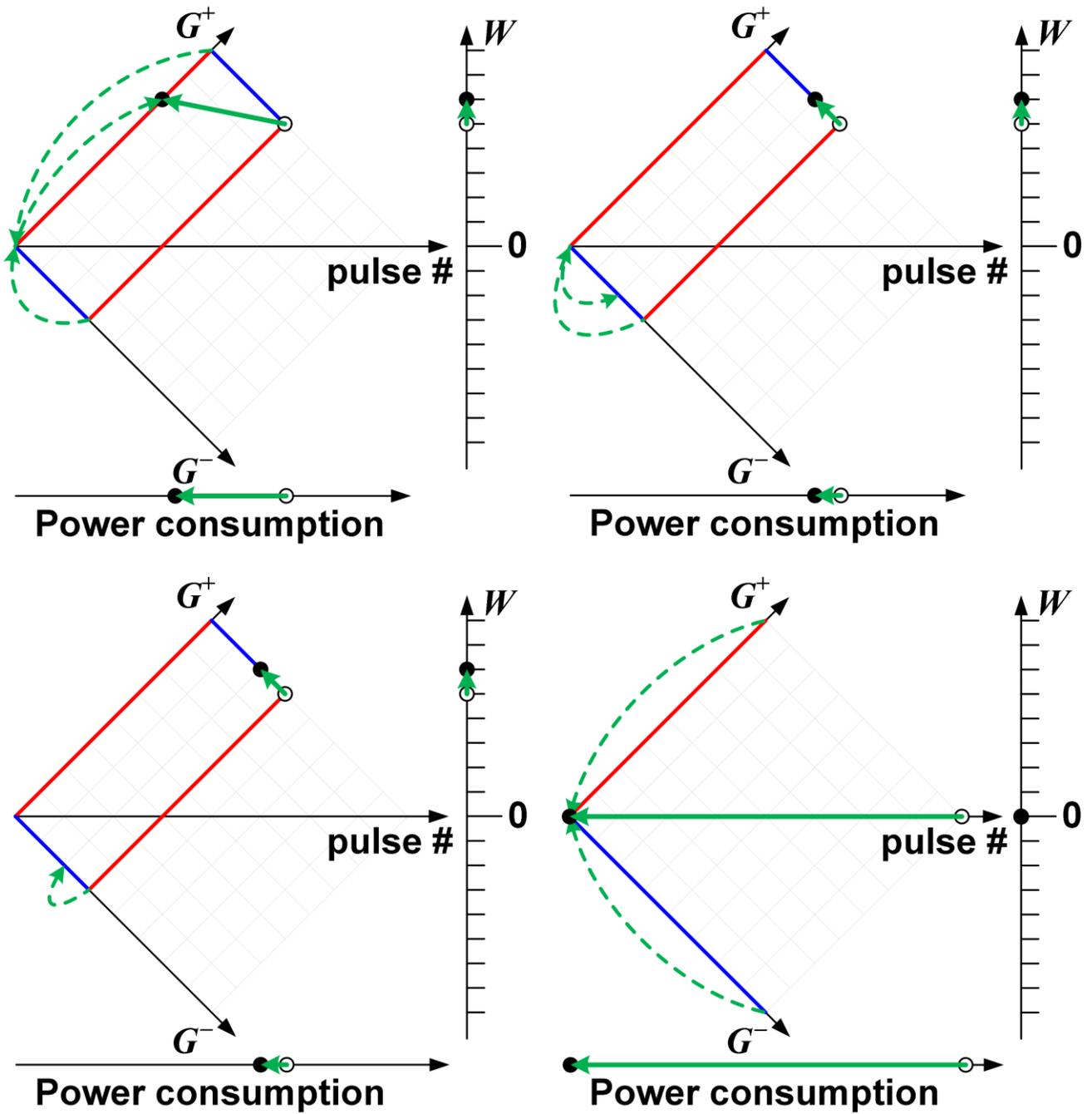

**Figure 2.** (**A**)–(**C**) Weight-updating methods when $G^+$ reaches $G_{\max}$: (**A**) reported unidirectional update method [25], (**B**) proposed unidirectional update method, (**C**) conventional bidirectional update method, and (**D**) initialization method when both $G^+$ and $G^-$ reach $G_{\max}$

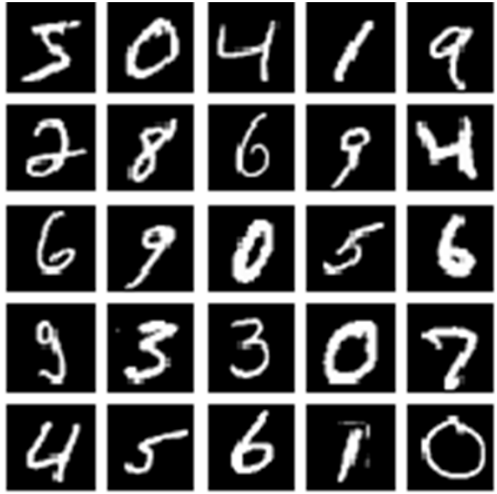
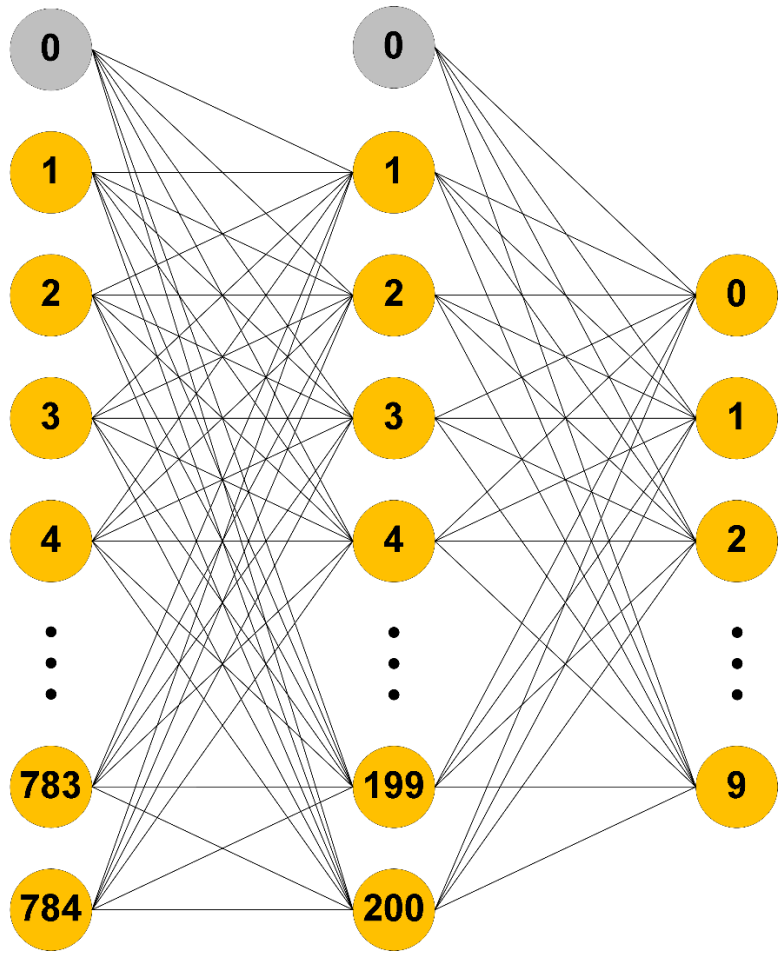

**Figure 3.** A three-layer perceptron network. Here, 784 input neurons are used for MNIST images (28×28), with 200 hidden neurons and ten output neurons to identify each digit. Each layer has one bias neuron.

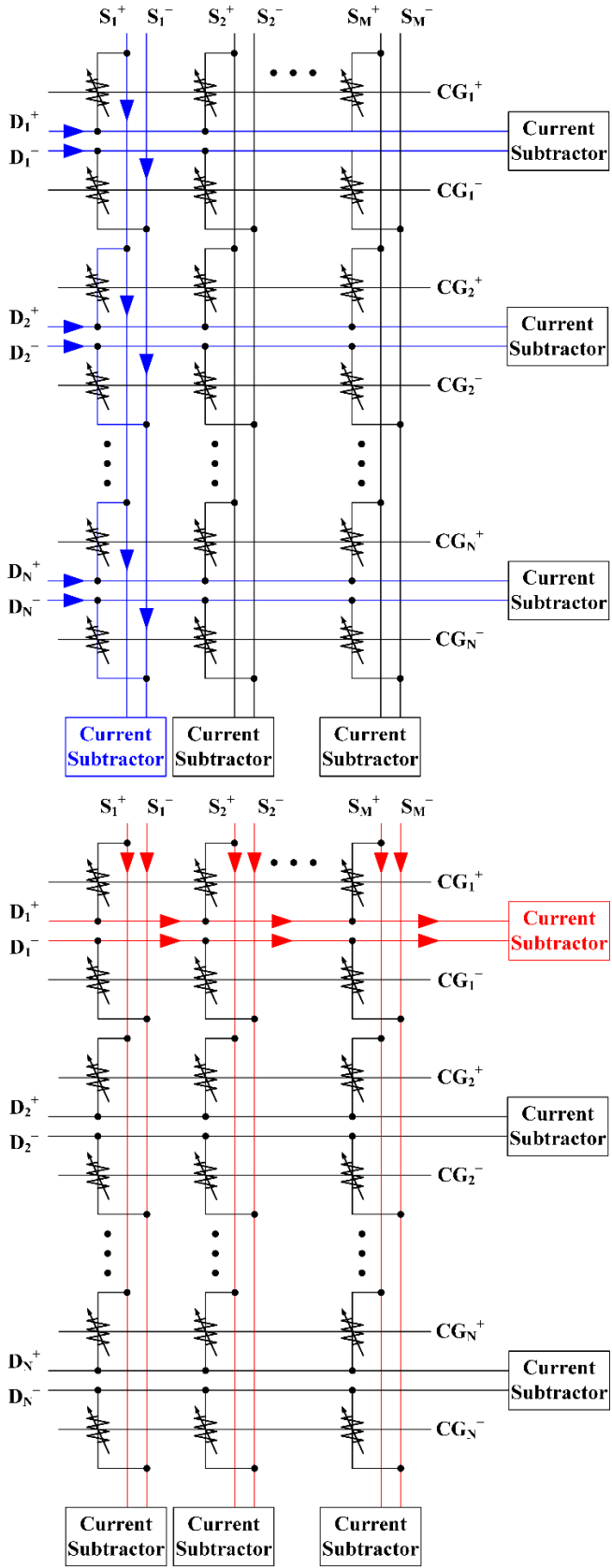

**Figure 4.** Implementation of vector-by-matrix multiplication for (**A**) forward propagation and (**B**) backward propagation using an electronic synapse device array.

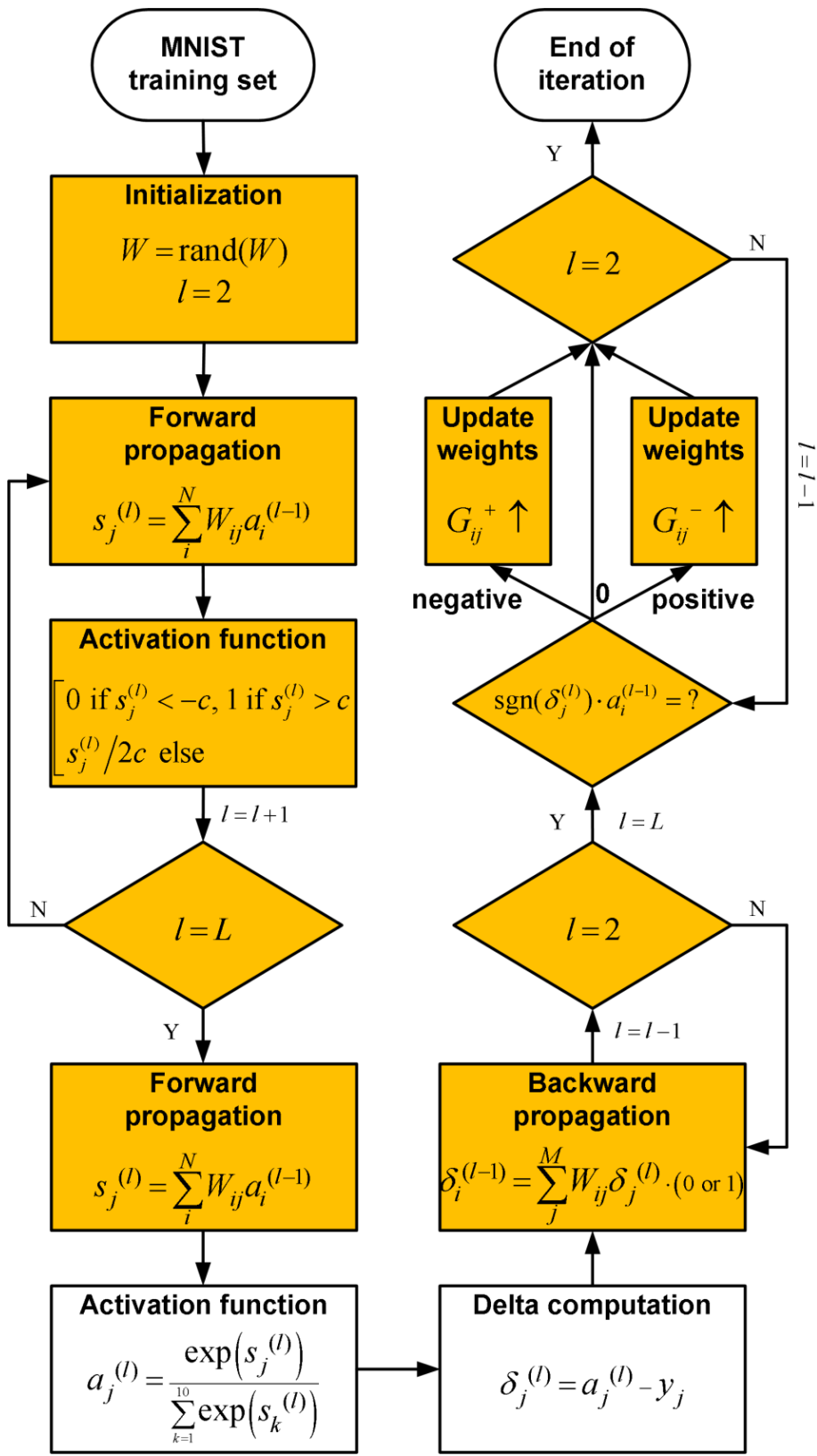

**Figure 5.** Online learning procedure. Shaded boxes are implemented in hardware. Here, '*L*' represents the number of layers.

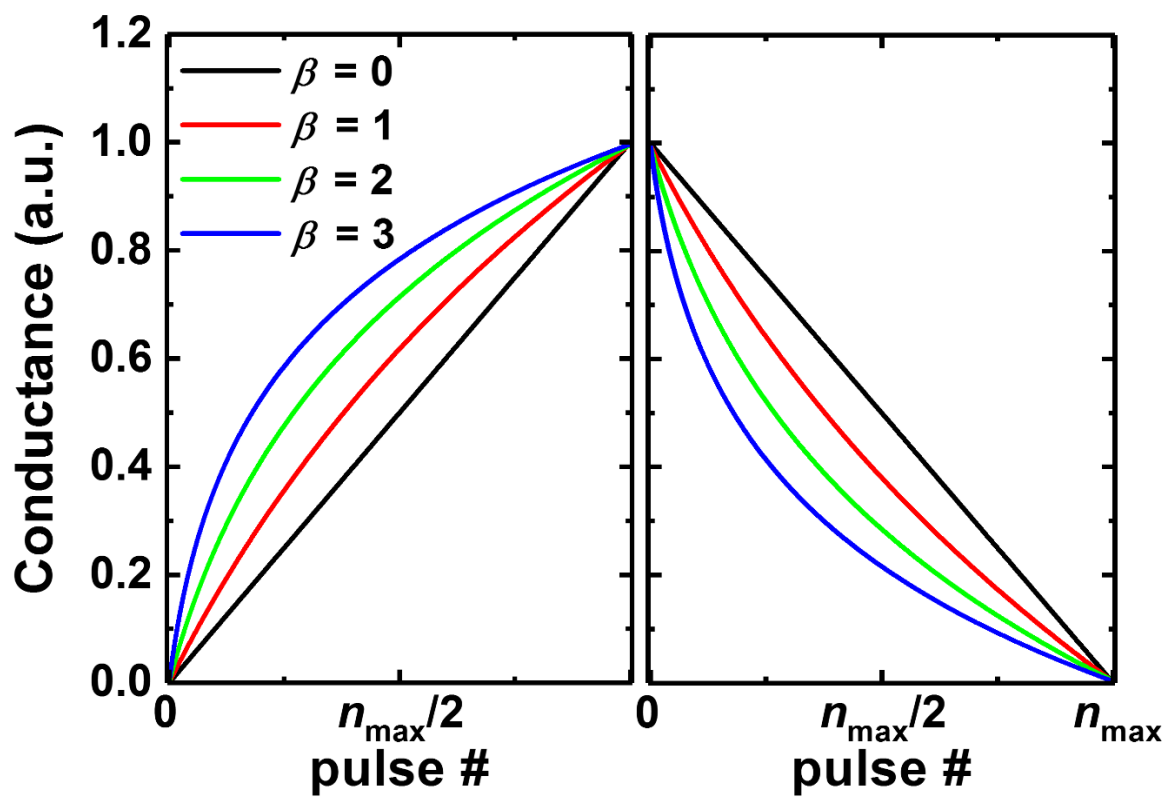

**Figure 6.** Potentiation and depression characteristics with respect to the parameter $\beta$.

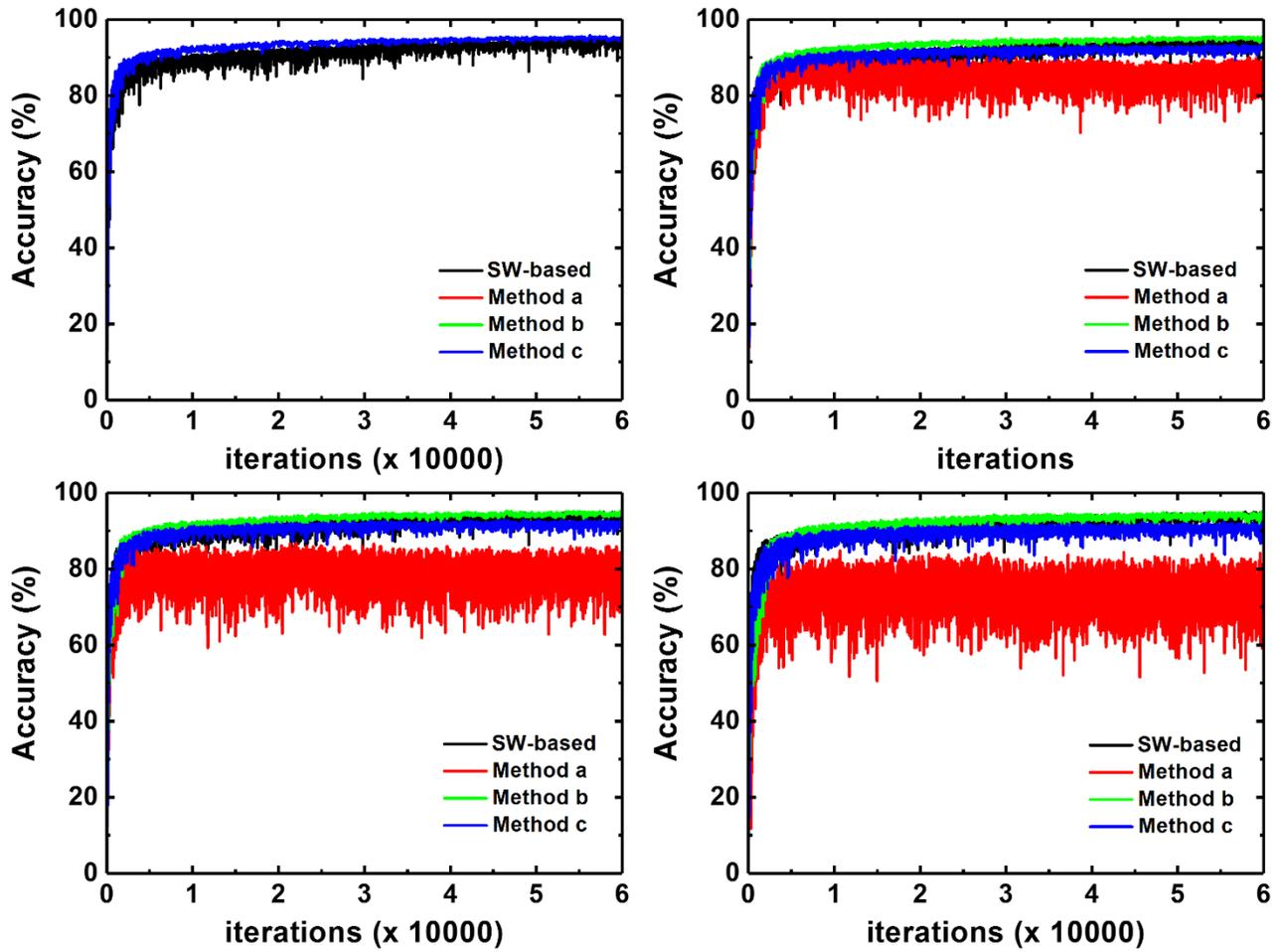

**Figure 7.** Classification accuracy with respect to the nonlinearity ($\beta$) and weight-updating methods. The nonlinearities ($\beta$) are **(A)** 0, **(B)** 1, **(C)** 2, and **(D)** 3.

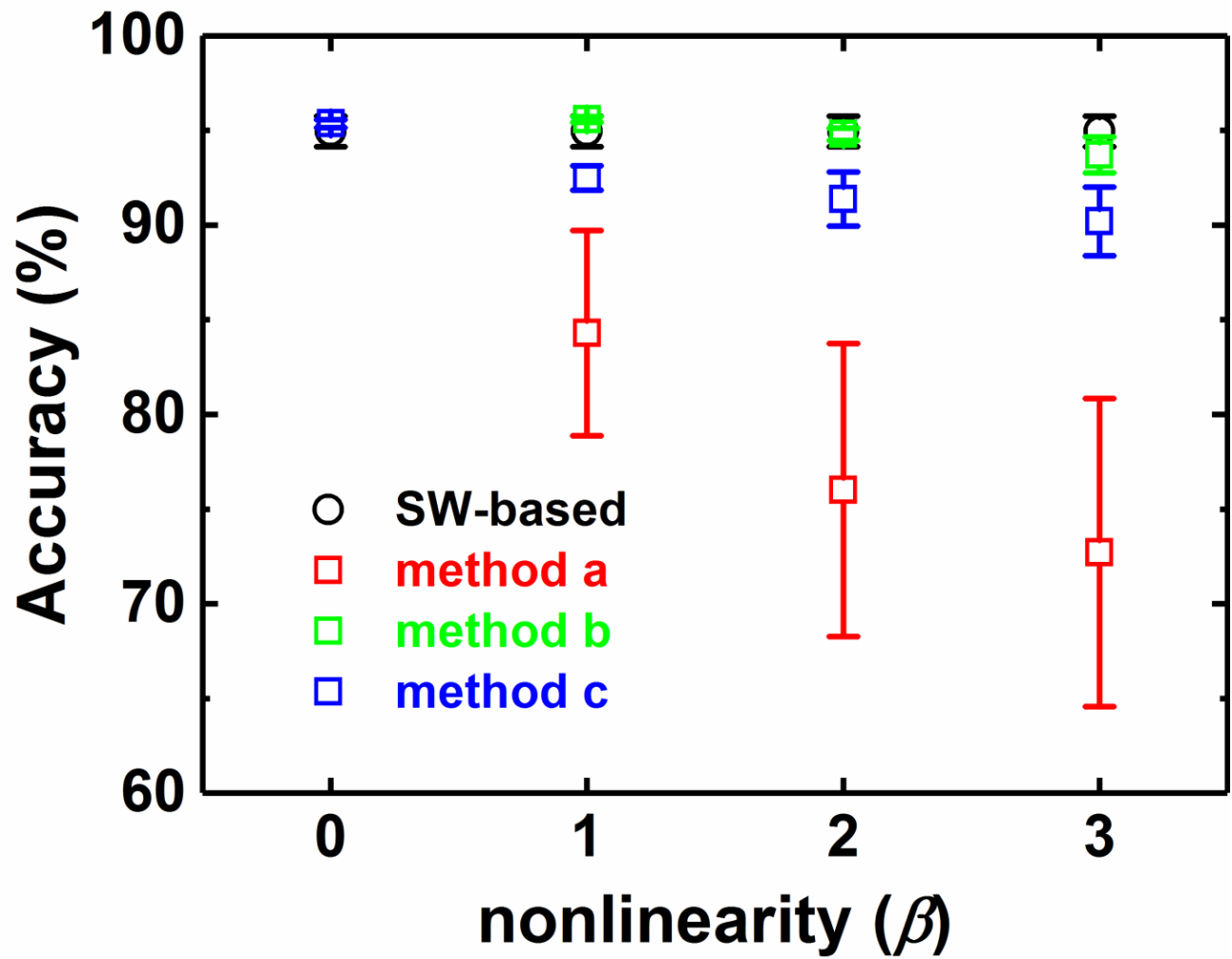

**Figure 8.** Classification accuracy with respect to the nonlinearity ($\beta$) and weight-updating methods when the dynamic range ($n_{max}$) is 64 and the mini-batch size is 1.

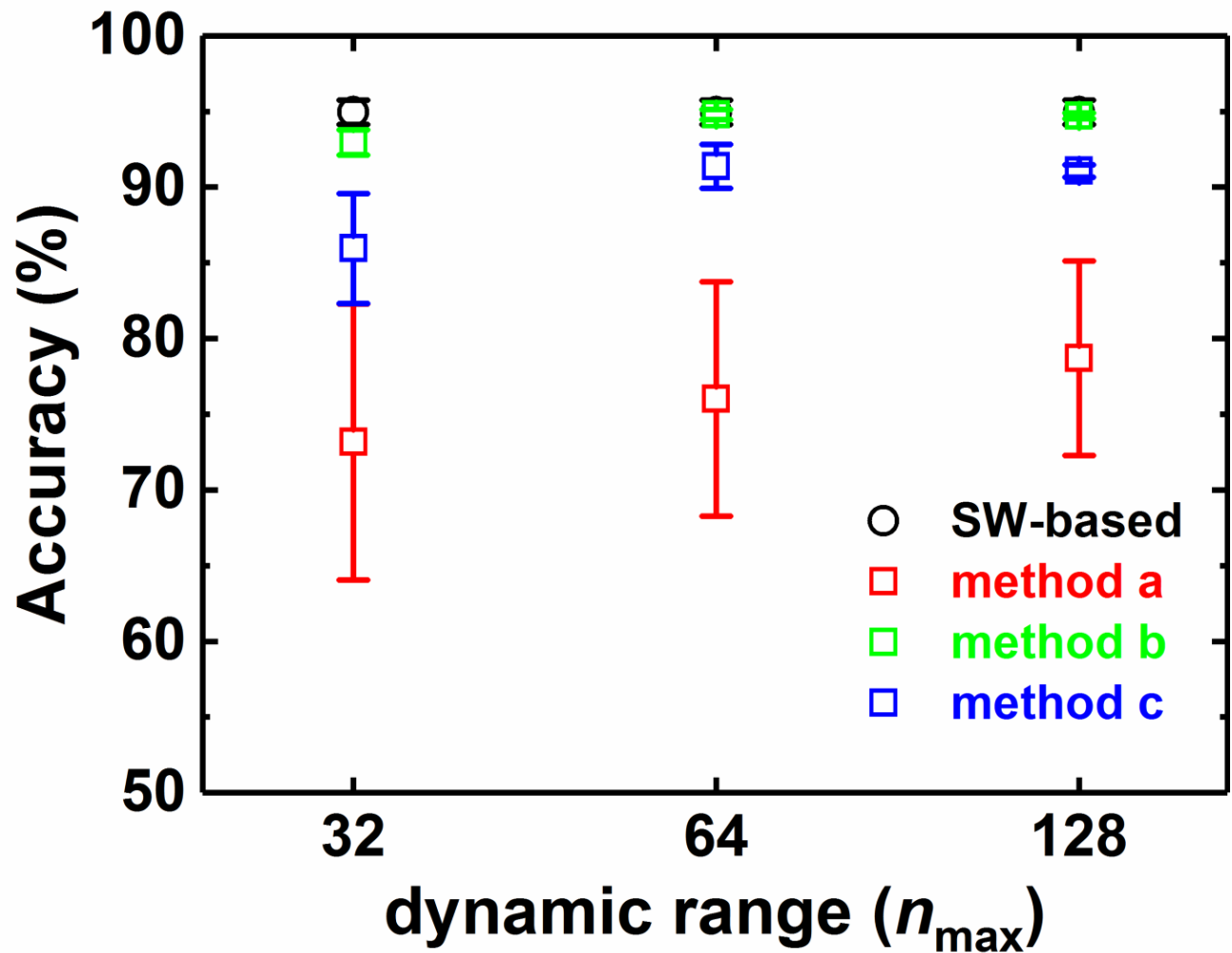

**Figure 9.** Classification accuracy with respect to the dynamic range ($n_{max}$) and weight-updating methods when the nonlinearity ($\beta$) is 2 and the mini-batch size is 1.

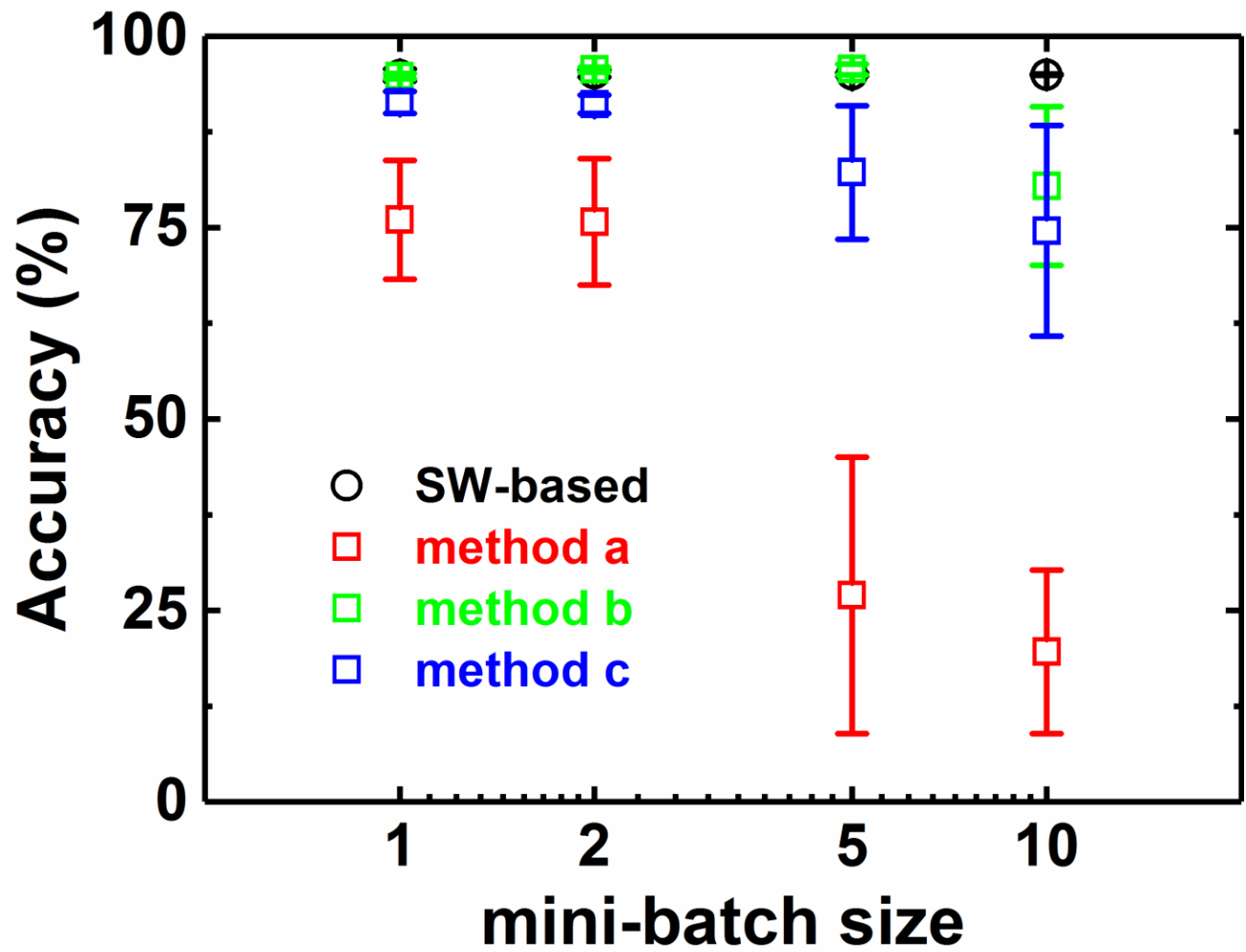

**Figure 10.** Classification accuracy with respect to the mini-batch size and weight-updating methods when the nonlinearity ($\beta$) is 2 and the dynamic range ($n_{max}$) is 64.

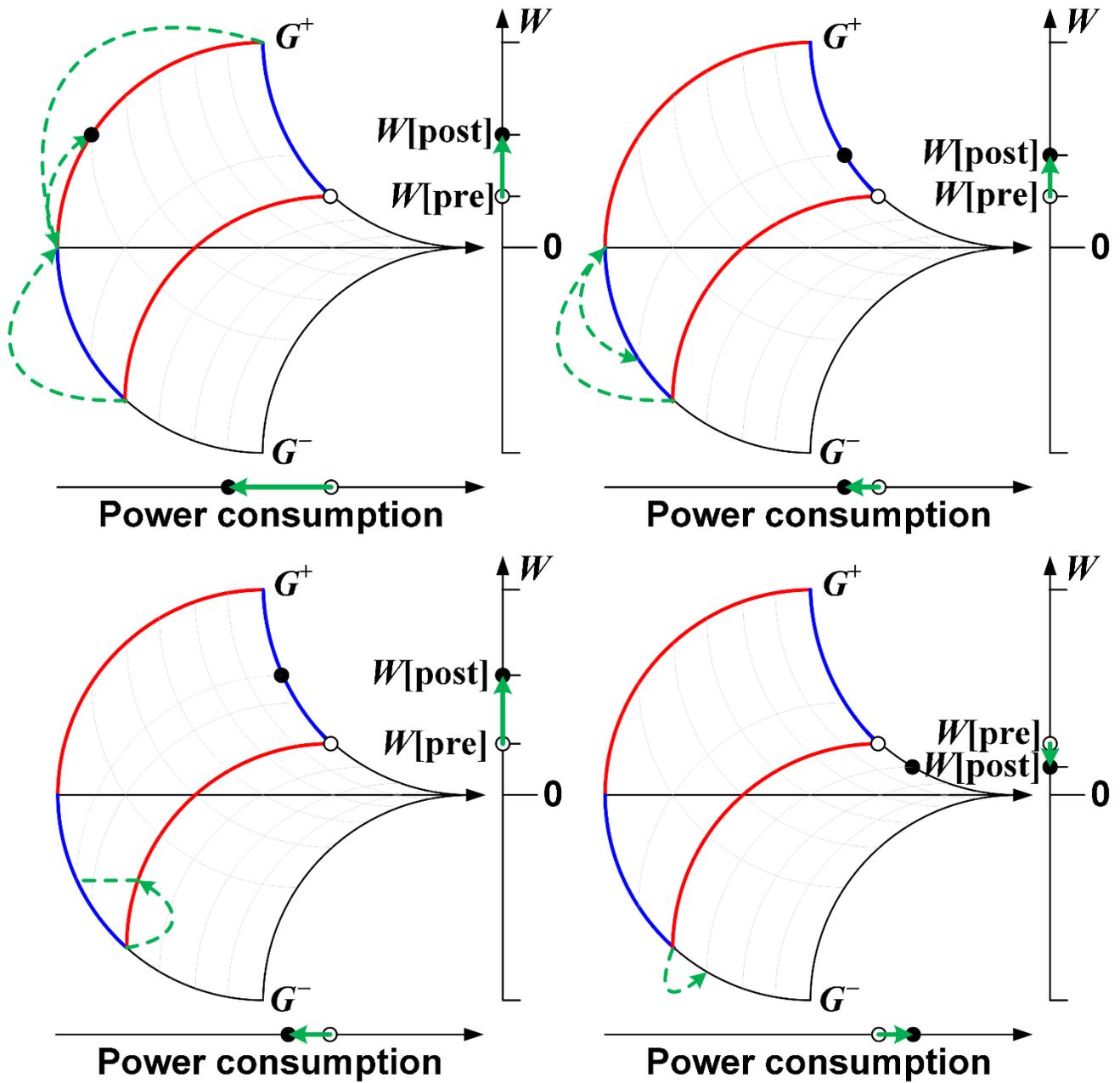

**Figure 11.** (**A**)–(**C**) Weight updates when $G^+ = G_{max}$ and $\Delta W > 0$ are given: (**A**) method a, (**B**) method b, and (**C**) method c. (**D**) Weight updates when $G^+ = G_{max}$ and $\Delta W < 0$ are given.

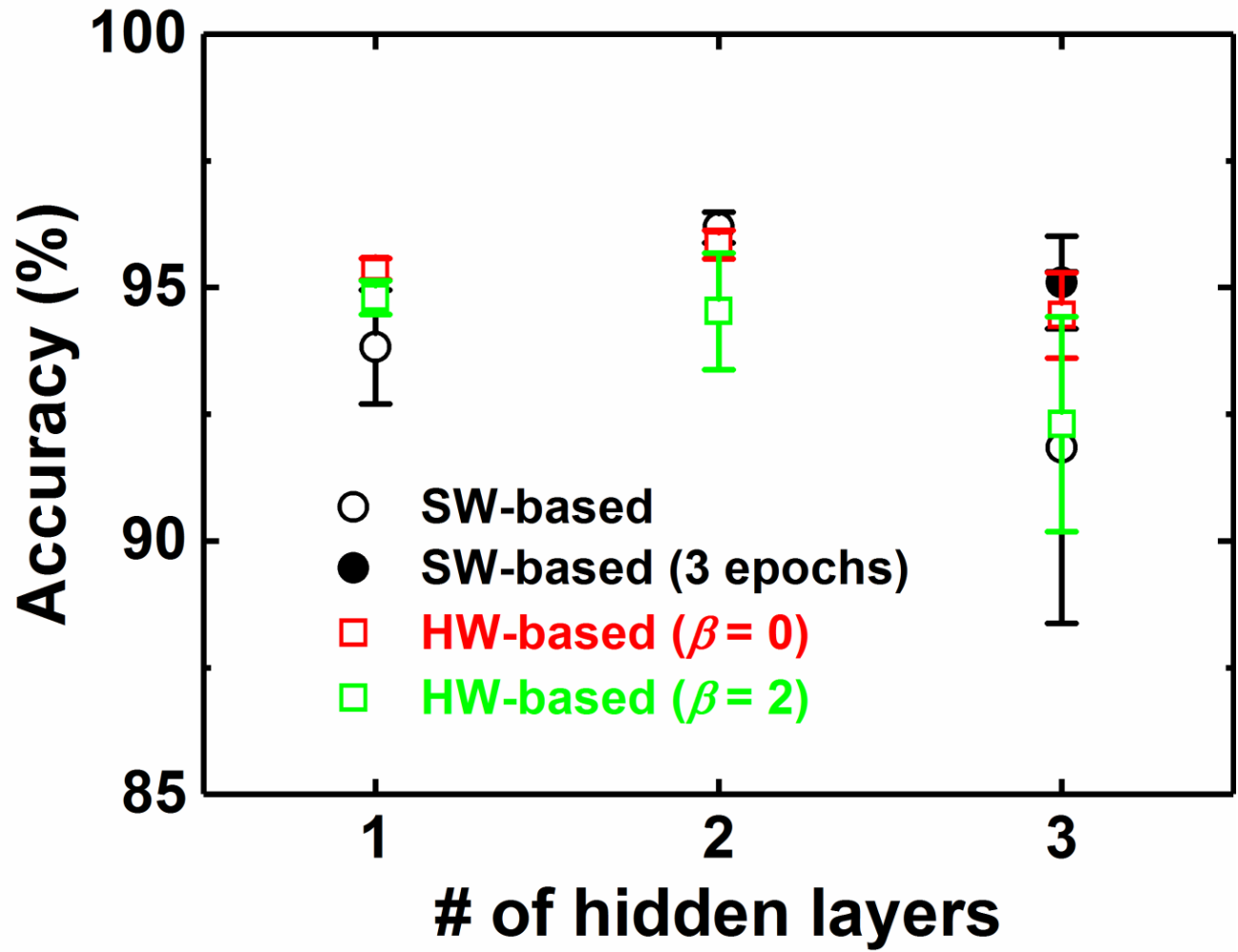

**Figure 12.** Classification accuracy with respect to the number of hidden layers when the nonlinearity ($\beta$) is 0, the dynamic range ($n_{max}$) is 64, the mini-batch size is 1, and method b is used. The accuracy rates for SW-based BP and HW-based BP ($\beta = 0, 2$) are represented by the black circle, and the red and green square symbols, respectively.

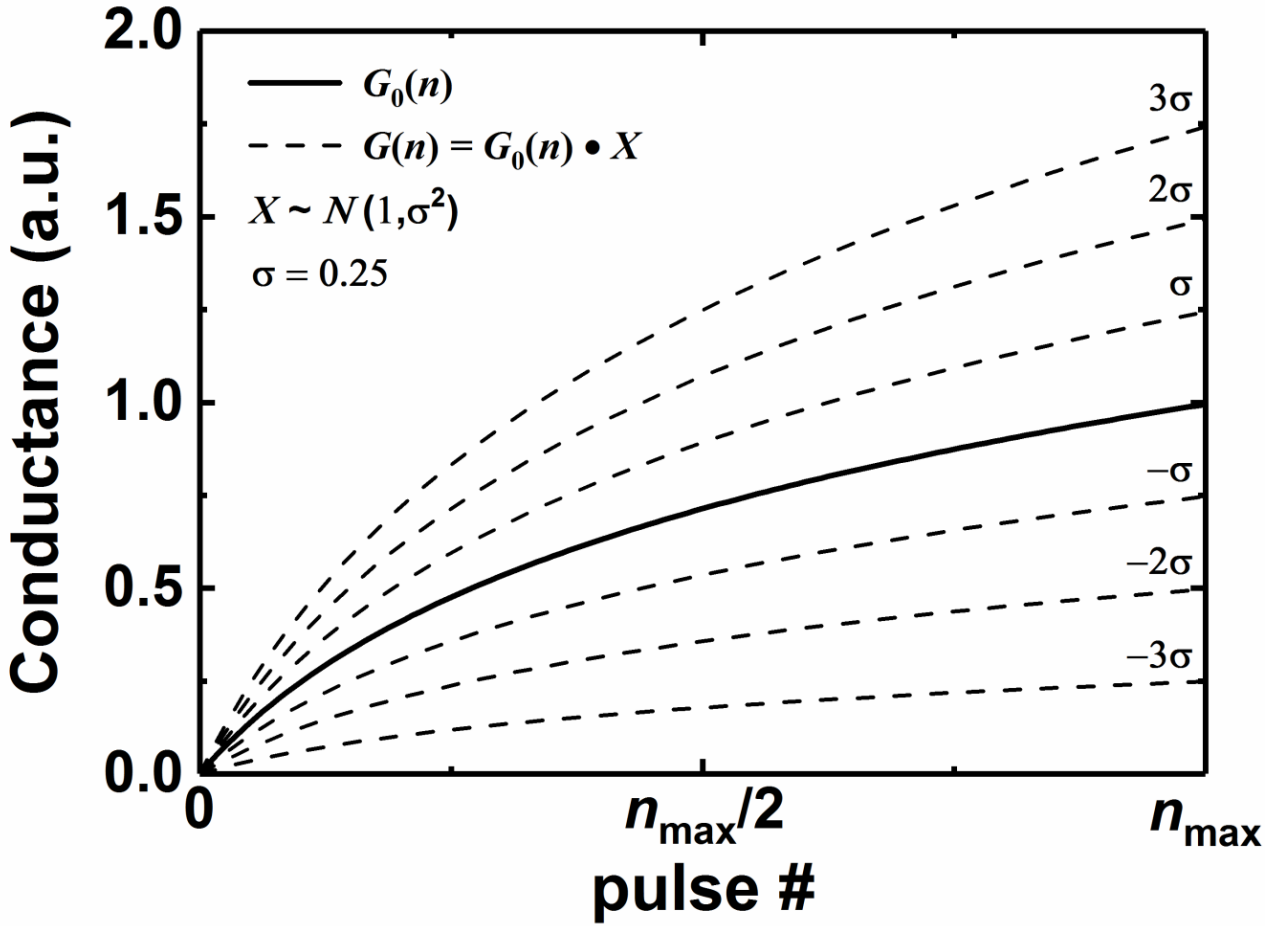

**Figure 13.** Relative conductance as a function of the number of pulses in cases with non-zero variance. For example, when the mean conductance is 1 and the standard deviation is ($\sigma$) 0.25, the conductance curves of $\pm 1\sigma, \pm 2\sigma,$ and $\pm 3\sigma$ are indicated by the dashed line.

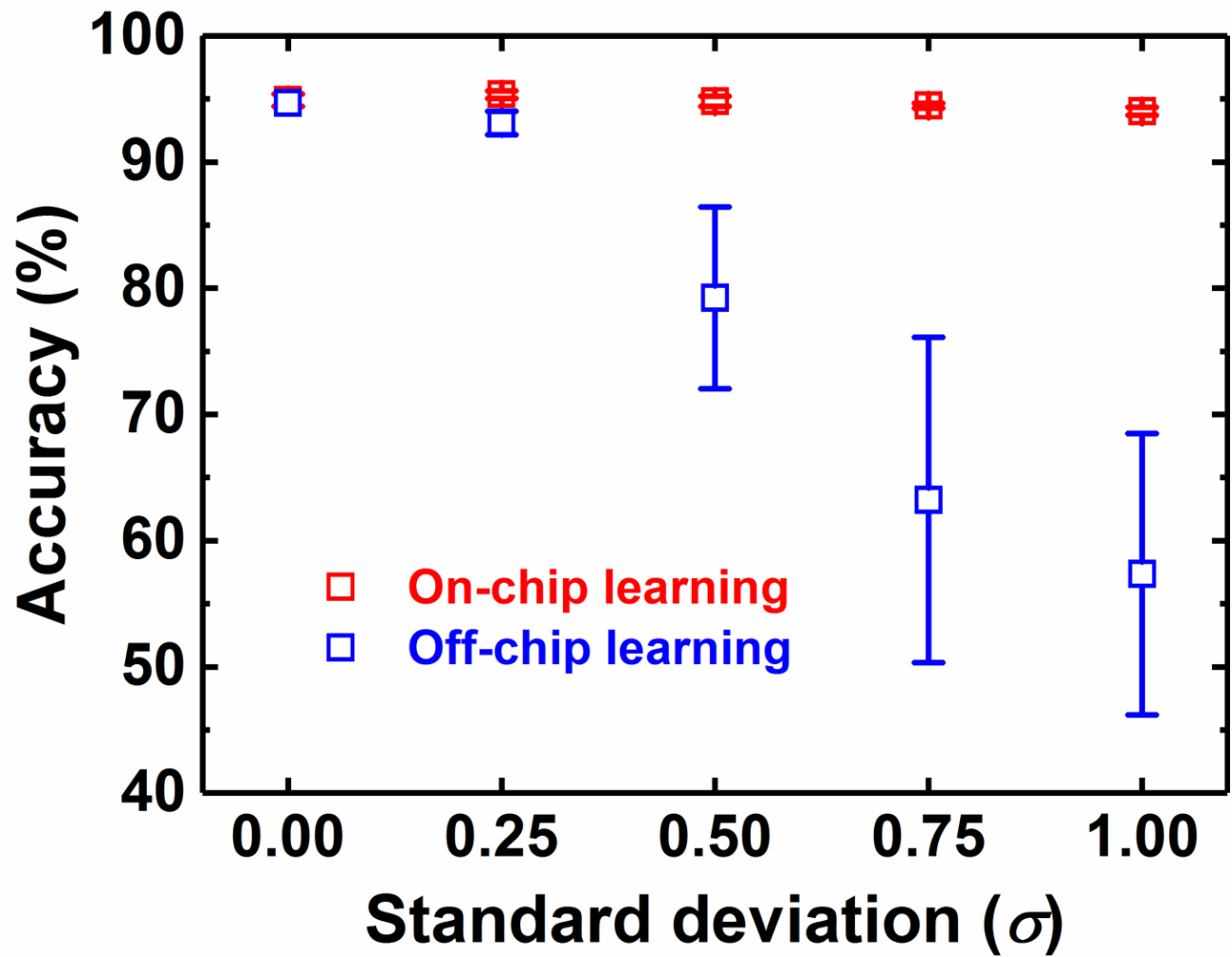

**Figure 14.** Classification accuracy with respect to the standard deviation (σ) when the nonlinearity ($\beta$) is 2, the dynamic range ($n_{max}$) is 64, and method b is used. The accuracy rates for on-chip learning and off-chip learning are represented by the red and blue square symbols, respectively.